\documentclass{article}

\usepackage[preprint]{neurips_2025}
\workshoptitle{Up-to-date Foundation Models: Continual Learning, Adaptation, and Evaluation}

\usepackage[utf8]{inputenc}
\usepackage[T1]{fontenc}
\usepackage{microtype}
\usepackage{xcolor}

\usepackage{amsmath, amsfonts}
\usepackage{booktabs}
\usepackage{nicefrac}

\usepackage{hyperref}
\usepackage{url}
\usepackage[pdftex]{graphicx}
\usepackage[inkscapelatex=false]{svg}

\newcommand{\ptpp}{\ensuremath{\mathit{P}\mathrm{\textsc{tpp}}}} 
\newcommand{\atpp}{\ensuremath{\mathit{A}\mathrm{\textsc{tpp}}}} 
\newcommand{\dcpt}{\ensuremath{\text{D}\text{-}\textsc{cpt}}}

\newcommand{\huberlog}{\ensuremath{\mathrm{Huber}_{\log}}}
\newcommand{\rmseLog}{\ensuremath{\mathrm{RMSE}_{\log}}}
\newcommand{\maerel}{\ensuremath{\mathrm{MAE}_{\mathrm{rel}}}}
\newcommand{\mapeclip}{\ensuremath{\mathrm{MAPE}_{\mathrm{clip}}}}

\title{PTPP-Aware Adaptation Scaling Laws:\\ Predicting Domain-Adaptation Performance at Unseen Pre-Training Budgets}

\author{%
  Etienne Goffinet \\
  Cerebras Systems \\
  \texttt{etienne.goffinet@cerebras.net} \\
  \And
  Shane Bergsma \\
  Cerebras Systems \\
  \texttt{shane.bergsma@cerebras.net} \\
  \And
  Avraham Sheinin \\
  Cerebras Systems \\
  \And 
  Natalia Vassilieva \\
  Cerebras Systems \\
  \And
  Shaheer Muhammad \\
  Cerebras Systems \\
  \And
  Preslav Nakov \\
  MBZUAI, Abu Dhabi \\
  Abu Dhabi, UAE \\
  \And
  Gurpreet Gosal \\
  Cerebras Systems \\
  \texttt{gurpreet.gosal@cerebras.net} \\
}

\begin{document}

\maketitle

\begin{abstract}
Continual pre-training (CPT) for domain adaptation must balance target-domain gains with stability on the base domain. Existing CPT scaling laws typically assume a fixed pre-training budget, which limits their ability to forecast adaptation outcomes for models trained at different tokens-per-parameter (PTPP). We present \emph{PTPP-aware} adaptation scaling laws that make the pre-training budget an explicit variable, enabling accurate \emph{prediction} of adaptation loss at unseen \ptpp. On a multilingual setup (English/Arabic $\rightarrow$ French), PTPP-aware formulations trained on early stages (\ptpp{}=\{15,31\}) predict target loss at \ptpp{}=279 and outperform a PTPP-agnostic \dcpt{} transfer baseline on metrics (Huber-on-log, MAE$_\mathrm{rel}$, calibration slope); full diagnostics (RMSE, MAPE) are in the appendix. Beyond forecasting, we show a practical use case: planning replay ratios and adaptation token budgets that satisfy target and forgetting constraints under compute limits.
\end{abstract}

\section{Introduction}
Capabilities of LLMs (large language models) continue to scale with model size, data size, and thus the total compute used for pre-training. Language models trained on a mixture of domains, dominated by web-scale corpora, yield general LLMs~\citep{biderman2023pythiasuiteanalyzinglarge, dey2023cerebrasgptopencomputeoptimallanguage, gemmateam2025gemma3technicalreport, olmo20252olmo2furious, yang2025qwen3technicalreport}. These generalist models may not perform well in tasks requiring specialized knowledge (e.g., in fields such as medicine, law, finance) or those requiring language capabilities beyond the dominant pre-training language. We must therefore \emph{adapt} these models to new, domain-specific, or target-language-specific data. 
This adaptation process presents a fundamental challenge: achieving strong performance in the target domain while preserving general capabilities (avoiding catastrophic forgetting~\citep{kirkpatrick2017overcoming}). 
Various strategies have been proposed to minimize forgetting~\citep{chen2025mofomomentumfilteredoptimizermitigating, ostapenko2022continuallearningfoundationmodels, biderman2024loralearnsforgets, ibrahim2024simplescalablestrategiescontinually, gupta2023continualpretraininglargelanguage}. 

Pre-training scaling laws are well established—e.g., relations between model/data size and performance in \citet{kaplan2020scalinglawsneurallanguage, hoffmann2022trainingcomputeoptimallargelanguage}—whereas CPT-specific laws are comparatively underexplored. D-CPT extends Chinchilla with \emph{replay} to study compute-optimal CPT at a fixed pre-training stage \citep{que2024d}, and forgetting laws quantify degradation on the pre-training domain at that stage \citep{bethune2025scalinglawsforgettingfinetuning}. 

However, most CPT scaling laws—D-CPT and forgetting laws included—assume a \emph{fixed} pre-training budget (a single \ptpp{} stage), which limits \emph{forecasting} across budgets. Prior work indicates that \ptpp{} modulates learning dynamics and downstream adaptation \citep{springer2025overtrainedlanguagemodelsharder, ash2020warm, lyle2023understanding, kumar2024scaling}. We therefore condition explicitly on \ptpp{}, yielding \emph{PTPP-aware} adaptation laws that predict target-domain loss at unseen \ptpp{} and clarify replay–stage interactions. Our central question: can a law fit at early stages (\ptpp=\{15,31\}) \emph{forecast} target validation loss at \ptpp=279?

Although our experiments focus on language adaptation, treating \ptpp{} as an explicit driver of adaptation dynamics is broadly applicable. Prior multilingual adaptation work underscores the need to mitigate and estimate forgetting while acquiring target competence~\citep{de_Vries_2021,fujii2024continualpretrainingcrosslingualllm,huang2024acegptlocalizinglargelanguage,gosal2024bilingualadaptationmonolingualfoundation,zhao2024llamaenglishempiricalstudy}. Our contributions include:
\begin{enumerate}
    \item \textbf{PTPP-aware adaptation laws.} We extend CPT scaling laws by integrating the pre-training budget (\ptpp) as an explicit variable in the functional form.
    \item \textbf{Forecasting at unseen \ptpp.} Fits at \ptpp{}={15,31} predict French loss at \ptpp{}=279 and outperform a \ptpp-agnostic \dcpt{} baseline on all metrics; a handful of 241M-scale “anchor” points at \ptpp{}=279 (20 calibration measurements at the evaluation stage) further improve accuracy at low-cost.
    \item \textbf{Planning under constraints.} Using the fitted law, we find an optimal replay ratio and adaptation token budget that satisfy target and forgetting constraints under compute limits.
\end{enumerate}

\section{Methodology and Experiments}
\paragraph{Setup.}
We study loss $\hat L$ as a function of model size $N$, adaptation tokens $D$, replay ratio $r\!\in(0,1]$ (s.t. $1-r$ is the target domain fraction), and pre-training $\ptpp$.
We use GPT-2–style decoder-only models pre-trained on a mixed English–Arabic corpus; the \emph{adaptation domain} is French. Fits use \ptpp{}=\{15,31\} and are \emph{evaluated} on \ptpp{}=279 (unseen), across $r\in\{0.10,0.25,0.50\}$ and $N\in\{$241M, 517M, 1.4B, 8.1B$\}$.

\paragraph{$\ptpp$-Aware candidate formulations (1–3).}
All laws share an $N$-term and an $r$-barrier; they differ in how \ptpp{} affects the data-efficiency term. Let $\varepsilon\!=\!10^{-5}$.

\paragraph{(1) Additive \ptpp{} prior (floor).}
\begin{align*}
\hat L \;=\; E \;+\; \frac{A}{N^{\alpha}}
\;+\; \frac{B\,r^{\nu}}{D^{\beta}}
\;+\; \frac{C}{(r+\varepsilon_r)^{\gamma}}
\;+\; \frac{F}{\ptpp^{\eta}} .
\end{align*}
\emph{\ptpp{} lowers the \emph{floor} of $\hat L$ via an additive term.}

\paragraph{(2) \ptpp\text{-}gated data exponent (no floor).}
\begin{align*}
\hat L \;=\; E \;+\; \frac{A}{N^{\alpha}}
\;+\; \frac{B\,r^{\nu}}{D^{\,\beta_{\!\text{eff}}}}
\;+\; \frac{C}{(r+\varepsilon_r)^{\gamma}},
\qquad
\beta_{\!\text{eff}}=\beta\!\left(1-\lambda\frac{\ptpp^{\zeta}}{1+\ptpp^{\zeta}}\right),\;\; \beta_{\!\text{eff}}\ge 10^{-6}.
\end{align*}
\emph{\ptpp{} controls the \emph{shape} of the data law via a bounded gate $\lambda\,\frac{\ptpp^{\zeta}}{1+\ptpp^{\zeta}}\in[0,\lambda)$, so $\beta_{\!\text{eff}}=\beta\,(1-g(\ptpp))$, representing the impact of pre-training budget on adaptation efficiency.}

\paragraph{(3) \ptpp\text{-}gated data exponent \emph{+} floor.}
\begin{align*}
\hat L \;=\; E \;+\; \frac{A}{N^{\alpha}}
\;+\; \frac{B\,r^{\nu}}{D^{\,\beta_{\!\text{eff}}}}
\;+\; \frac{C}{(r+\varepsilon_r)^{\gamma}}
\;+\; \frac{F}{\ptpp^{\eta}},
\qquad
\beta_{\!\text{eff}}=\beta\!\left(1-\lambda\frac{\ptpp^{\zeta}}{1+\ptpp^{\zeta}}\right).
\end{align*}
\emph{\ptpp{} acts \emph{twice}: (i) a bounded gate reshapes the $D$-response (as in Form~2), and (ii) an additive prior $F/\ptpp^\eta$ lowers the loss floor. This captures both shape and offset effects of pre-training.}

\paragraph{Anchors.}
We also report a few-shot variant that augments the fit with 20 small-scale (241M) \emph{anchors} collected at the evaluation stage (\ptpp{}=279) across the $(r,D)$ grid; all other \ptpp{}=279 points remain held out (unlike the oracle, which fits on the full \ptpp{}=279 set). These anchors tighten calibration (slope $\to 1$) and error metrics at low-cost.

\paragraph{Data \& models.}
GPT-2–style decoders are pre-trained on English/Arabic (source) and adapted to French (target). We consider \ptpp{}$\in\{15,31,279\}$; replay $r\in\{0.10,0.25,0.50\}$; and model sizes $\{$241M, 517M, 1.4B, 8.1B$\}$. We focus on the French target; source-domain (English/Arabic) results are deferred to the appendix.

\paragraph{Fitting constraints.}
We minimize Huber loss on log residuals ($\delta{=}0.02$) with L\textendash BFGS\textendash B under positivity constraint for all parameters except $\zeta\in\mathbb{R}$. We clip $r\!\in[10^{-9},1-10^{-9}]$.

\paragraph{Metrics.}
We assess predictions at \ptpp=279 with three metrics. \emph{Huber-on-log} is the Huber loss to residuals $r=\log\hat y-\log y$ with $\delta=0.02$. \emph{MAE$_{\text{rel}}$} is the mean absolute relative error $\tfrac{1}{n}\!\sum_i \tfrac{|\hat y_i-y_i|}{y_i}$, i.e., the typical percentage miss (lower is better). \emph{Calibration (intercept/slope)}: parameters $(a,b)$ from an Ordinary Least Squares (OLS) fit $\log y = a + b\,\log \hat y$; ideal is $a\approx 0$, $b\approx 1$.

\section{Results: Forecasting at Unseen \ptpp}

\begin{table}[h!]
\centering
\small
\begin{tabular}{lccc}
\toprule
Formulation & \(\huberlog\!\downarrow\) & \(\maerel\!\downarrow\) & Calib.\ slope \(\approx 1\) \\
\midrule
Form 1 (Additive Prior)     & \(2.34\!\times\!10^{-4}\) & \(2.08\!\times\!10^{-2}\) & \(0.991\) \\
Form 2 (Gated Exponent)     & \(1.99\!\times\!10^{-4}\) & \(1.83\!\times\!10^{-2}\) & \(0.970\) \\
\textbf{Form 3 (Gated+Floor)} & \textbf{\(4.43\!\times\!10^{-5}\)} & \textbf{\(6.70\!\times\!10^{-3}\)} & \textbf{\(0.991\)} \\
\midrule
\dcpt{} (no \ptpp, transfer) & \(4.74\!\times\!10^{-4}\) & \(3.43\!\times\!10^{-2}\) & \(0.961\) \\
\bottomrule
\end{tabular}
\caption{French prediction at unseen \ptpp{}=279 (no anchors; trained on \ptpp{}=\{15,31\}). 
Full metrics (including calibration intercepts and RMSE) appear in Appendix~\ref{app:full-metrics}.}
\label{tab:pred_noanchors}
\end{table}

\begin{figure}[h!]
    \centering
    \includegraphics[width=\linewidth]{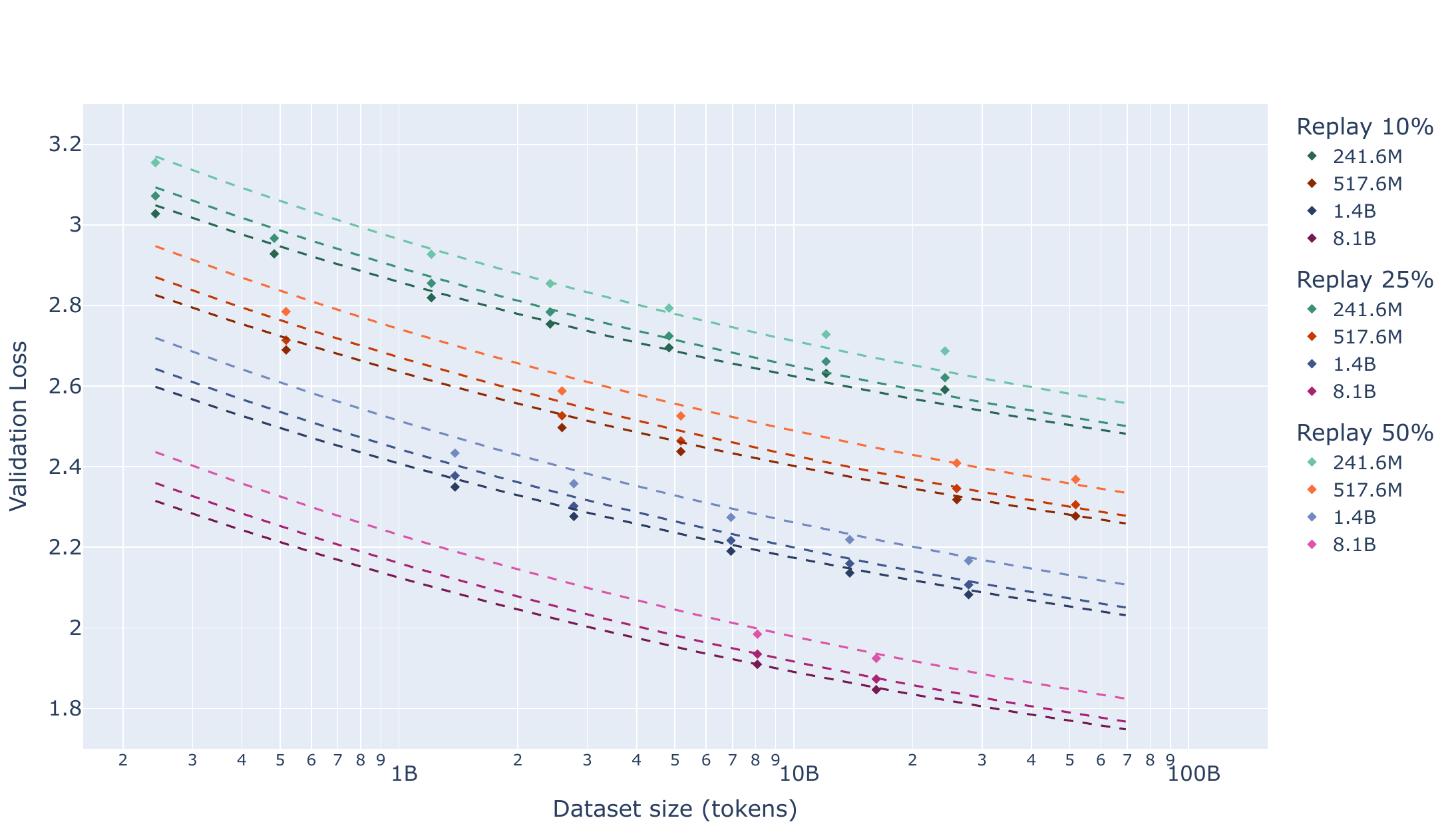}
    \caption{\ptpp=279 \emph{predictions} of the \emph{gated+floor} model (dashed) vs.\ \emph{observations} (markers) of validation loss for $r\in\{0.10,0.25,0.50\}$ and \{241M, 517M, 1.4B, 8.1B\}.}
    \label{fig:ptpp_aware_predictions}
\end{figure}

\begin{table}[t]
\centering
\small
\begin{tabular}{lccc}
\toprule
Formulation & \(\huberlog\!\downarrow\) & \(\maerel\!\downarrow\) & Calib.\ slope \(\approx 1\) \\
\midrule
Form 1 (Additive Prior)     & \(5.22\!\times\!10^{-5}\) & \(8.56\!\times\!10^{-3}\) & \(0.956\) \\
Form 2 (Gated Exponent)     & \(4.23\!\times\!10^{-5}\) & \(8.17\!\times\!10^{-3}\) & \(0.992\) \\
\textbf{Form 3 (Gated+Floor)} & \textbf{\(3.54\!\times\!10^{-5}\)} & \textbf{\(7.39\!\times\!10^{-3}\)} & \textbf{\(0.992\)} \\
\bottomrule
\end{tabular}
\caption{French prediction at unseen \ptpp{}=279 with 20 anchors at 241M-scale.}
\label{tab:pred_anchors}
\end{table}

\paragraph{Takeaways.}
Across French at unseen \ptpp{}=279, the \emph{gated+floor} variant (Form 3) is consistently best, with low errors and near-ideal calibration (slope ${\approx}0.99$) both without anchors and with 20 small-scale anchors; the \emph{gated-only} variant (Form 2) is reliably second and ahead of \dcpt{} transfer. Anchors uniformly tighten Huber/RMSE and calibration without changing the methods' rankings.

On the English/Arabic source domain (Appendix~\ref{app:full-metrics}), the picture depends on supervision at the evaluation stage: without anchors, \emph{floor-only} (Form 1) suffices—suggesting \ptpp{} mainly shifts the baseline—whereas with anchors the \emph{gated-only} form (Form 2) becomes best, revealing a data-efficiency (shape) effect once lightly calibrated at \ptpp{}=279. Overall, results support that a) the preferred functional form can be domain and/or supervision-dependent and b) a direct link exists between pre-training compute and adaptation efficiency that manifests as both a floor shift and a learning-curve shape change; few-shot anchors prove to be a low-cost way to calibrate the latter.

\section{Use Case: Joint Compute and Replay Optimization}

\begin{figure}
    \centering
    \includegraphics[width=1.0\linewidth]{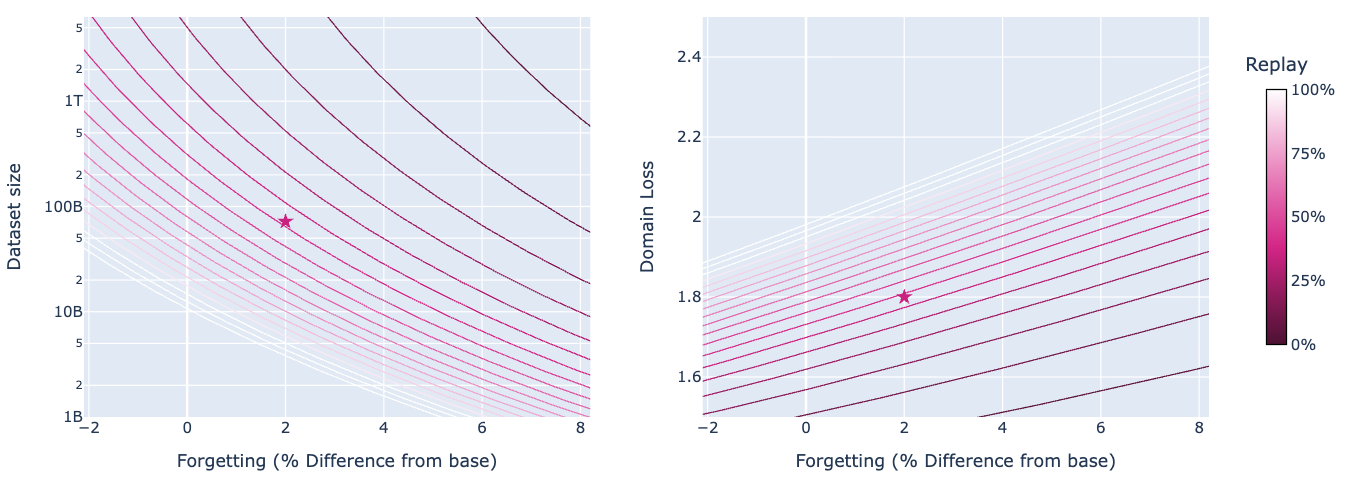}
    \caption{Replay (0–100\%) determines the trade-off between forgetting and domain performance.
    Left: Forgetting / Dataset size landscape.
    Right: resulting French loss. The star highlights the solution (8.9 $\atpp$, 34\% replay), minimizing FLOPs s.t. forgetting is $\leq$+2\% and French loss $\leq$ 1.8.}
    \label{fig:use_case}
\end{figure}

In domain adaptation, one must balance \emph{forgetting} of the source domain with improvements in the target domain, under strict compute budgets. A key feature of our method is that ptpp-aware scaling-law fits allow prediction of both losses at an unseen $\ptpp$ (279), making it possible to solve this trade-off analytically rather than through brute-force sweeps. 
We consider a target model scale of $N=8.1$B, pretrained at $\ptpp=279$, and seek the smallest adaptation tokens-per-parameter (\atpp{}) that meets the forgetting and target-performance constraints, under the Form 1 hypothesis for English/Arabic loss and Form 3 for French. Let the adaptation budget $\atpp=D/N$. We solve:
\[
\min_{\atpp\ge 0,\; r\in[0,1]} \;\; \atpp
\quad \text{s.t.}\quad
\Delta L_{\mathrm{src}}\bigl(N,\; D,\; r,\; 279\bigr) \le \delta,\;\;
L_{\mathrm{tgt}}\bigl(N,\; D,\; r,\; 279\bigr) \le \tau .
\]

where $\Delta L_{\mathrm{src}} = L_{\mathrm{src}}(N,D,r,\ptpp)-L_{\mathrm{src}}(N,0,1,\ptpp)$, $N$ is model size, and $r\in[0,1]$ the replay ratio. Constraints are given by tolerated forgetting \(\delta\) (e.g.\ +2\%) and target French loss threshold \(\tau\)=1.8.  The optimal solution, displayed on Fig.~\ref{fig:use_case}, is \(\atpp=8.9\) and replay 34\%.

\section{Conclusion}

We proposed \emph{PTPP-aware} adaptation scaling laws that condition on the pre-training budget and \emph{predict} target performance at unseen $\ptpp$. On French at $\ptpp{=}279$, laws fit at early stages ($\ptpp\in\{15,31\}$) generalize well and outperform a $\ptpp$-agnostic \dcpt{} transfer baseline; a small set of 241M-scale anchors further improves accuracy. Empirically, pre-training progress modulates both the \emph{loss floor} and the adaptation efficiency, and a few low-cost anchors further enhance the prediction performances; on the source domain, floor shifts explain most gains without anchors, while the light anchoring reveals a data-dependent effect at $\ptpp{=}279$. These fits enable the optimization of replay and adaptation tokens under compute constraints. Promising directions include investigating how language transfer shapes the $\ptpp$ effect, extending to additional $\ptpp$ stages and domains, assessing task-level metrics, and adding uncertainty quantification with cost-aware anchor selection.

\bibliographystyle{plainnat}
\bibliography{references}

\appendix

\section*{Appendix A: Full metric tables}
\label{app:full-metrics}

\noindent\textbf{Metrics.}
Let $(y_i,\hat y_i)$ be observed and predicted \emph{validation losses} at the held-out stage (\ptpp{}=279). We evaluate errors primarily in \emph{log space} to capture multiplicative miss and stabilize heteroscedasticity. Define the log–residuals $r_i=\log\hat y_i-\log y_i$.

\emph{\huberlog} (\,$\downarrow$\,) is the mean Huber loss applied to $r_i$ with threshold $\delta{=}0.02$:
\[
\mathrm{Huber}_\delta(r)=
\begin{cases}
\frac{1}{2}r^2, & |r|\le\delta,\\[2pt]
\delta\bigl(|r|-\tfrac{1}{2}\delta\bigr), & |r|>\delta,
\end{cases}
\qquad
\huberlog=\frac{1}{n}\sum_i \mathrm{Huber}_\delta(r_i).
\]
It is quadratic near zero (like MSE) but linear for outliers, making it robust.

\emph{\rmseLog} (\,$\downarrow$\,) is the root-mean-square of the log–residuals,
\[
\rmseLog=\sqrt{\tfrac{1}{n}\sum_i r_i^2},
\]
which measures typical multiplicative error (e.g., $\rmseLog\!=\!0.01$ corresponds to $\approx\!1\%$ relative miss under small-error linearization).

\emph{\maerel} (\,$\downarrow$\,) is the mean absolute \emph{relative} error in the original scale,
\[
\maerel=\frac{1}{n}\sum_i \frac{|\hat y_i-y_i|}{y_i},
\]
i.e., the average percentage miss.

\emph{\mapeclip} (\,$\downarrow$\,) is a clipped MAPE that avoids division by tiny $y_i$:
\[
\mapeclip=\frac{1}{n}\sum_i \frac{|\hat y_i-y_i|}{\max(y_i,\,y_{\mathrm{clip}})},
\]
with a small $y_{\mathrm{clip}}>0$; when all $y_i\!\gg\!y_{\mathrm{clip}}$, \mapeclip{} equals \maerel.

\emph{Intercept/Slope} report calibration from the OLS fit
\[
\log y_i \;=\; a \;+\; b\,\log \hat y_i \;+\; \varepsilon_i.
\]
Perfect calibration gives $a{\approx}0$ (no systematic bias) and $b{\approx}1$ (correct sensitivity). We therefore seek small $|a|$ and $b$ close to $1$.

\paragraph{French — Unseen \ptpp=279, no anchors.}
\begin{center}
\small
\begin{tabular}{lcccccc}
\toprule
Formulation & Huber$_{\log}\downarrow$ & RMSE$_{\log}\downarrow$ & MAE$_{\text{rel}}\downarrow$ & MAPE$_{\text{clip}}\downarrow$ & Interc. & Slope \\
\midrule
Form 1 (Additive)       & $2.34\!\times\!10^{-4}$ & $2.27\!\times\!10^{-2}$ & $2.08\!\times\!10^{-2}$ & $2.08\!\times\!10^{-2}$ & $-0.01$ & $0.991$ \\
Form 2 (Gated)          & $1.99\!\times\!10^{-4}$ & $2.12\!\times\!10^{-2}$ & $1.83\!\times\!10^{-2}$ & $1.83\!\times\!10^{-2}$ & $0.05$  & $0.970$ \\
\textbf{Form 3 (G+F)}   & \textbf{$4.43\!\times\!10^{-5}$} & \textbf{$9.53\!\times\!10^{-3}$} & \textbf{$6.70\!\times\!10^{-3}$} & \textbf{$6.70\!\times\!10^{-3}$} & $0.01$  & $0.991$ \\
\midrule
\dcpt{} (transfer)      & $4.74\!\times\!10^{-4}$ & $3.47\!\times\!10^{-2}$ & $3.43\!\times\!10^{-2}$ & $3.43\!\times\!10^{-2}$ & $-0.00$ & $0.961$ \\
\bottomrule
\end{tabular}
\end{center}

\paragraph{French — Unseen \ptpp=279, with 241M anchors.}
\begin{center}
\small
\begin{tabular}{lcccccc}
\toprule
Formulation & Huber$_{\log}\downarrow$ & RMSE$_{\log}\downarrow$ & MAE$_{\text{rel}}\downarrow$ & MAPE$_{\text{clip}}\downarrow$ & Interc. & Slope \\
\midrule
Form 1 (Additive)       & $5.22\!\times\!10^{-5}$ & $1.02\!\times\!10^{-2}$ & $8.56\!\times\!10^{-3}$ & $8.56\!\times\!10^{-3}$ & $0.03$ & $0.956$ \\
Form 2 (Gated)          & $4.23\!\times\!10^{-5}$ & $9.20\!\times\!10^{-3}$ & $8.17\!\times\!10^{-3}$ & $8.17\!\times\!10^{-3}$ & $0.00$ & $0.992$ \\
\textbf{Form 3 (G+F)}   & \textbf{$3.54\!\times\!10^{-5}$} & \textbf{$8.42\!\times\!10^{-3}$} & \textbf{$7.39\!\times\!10^{-3}$} & \textbf{$7.39\!\times\!10^{-3}$} & $0.00$ & $0.992$ \\
\bottomrule
\end{tabular}
\end{center}

\paragraph{English/Arabic source — Unseen \ptpp=279, no anchors.}
\begin{center}
\small
\begin{tabular}{lcccccc}
\toprule
Formulation & Huber$_{\log}\downarrow$ & RMSE$_{\log}\downarrow$ & MAE$_{\text{rel}}\downarrow$ & MAPE$_{\text{clip}}\downarrow$ & Interc. & Slope \\
\midrule
\textbf{Form 1 (Additive)} & \textbf{$9.89\!\times\!10^{-5}$} & \textbf{$1.44\!\times\!10^{-2}$} & \textbf{$1.18\!\times\!10^{-2}$} & \textbf{$1.18\!\times\!10^{-2}$} & $-0.05$ & $1.034$ \\
Form 2 (Gated)          & $2.79\!\times\!10^{-4}$ & $2.75\!\times\!10^{-2}$ & $2.27\!\times\!10^{-2}$ & $2.27\!\times\!10^{-2}$ & $-0.03$ & $1.045$ \\
Form 3 (G+F)            & $7.55\!\times\!10^{-4}$ & $5.06\!\times\!10^{-2}$ & $4.65\!\times\!10^{-2}$ & $4.65\!\times\!10^{-2}$ & $0.01$  & $1.030$ \\
\midrule
\dcpt{} (transfer)      & $5.73\!\times\!10^{-4}$ & $4.08\!\times\!10^{-2}$ & $3.91\!\times\!10^{-2}$ & $3.91\!\times\!10^{-2}$ & $0.04$ & $0.932$ \\
\bottomrule
\end{tabular}
\end{center}

\paragraph{English/Arabic source — Unseen \ptpp=279, with 241M anchors.}
\begin{center}
\small
\begin{tabular}{lcccccc}
\toprule
Formulation & Huber$_{\log}\downarrow$ & RMSE$_{\log}\downarrow$ & MAE$_{\text{rel}}\downarrow$ & MAPE$_{\text{clip}}\downarrow$ & Interc. & Slope \\
\midrule
Form 1 (Additive)       & $9.21\!\times\!10^{-5}$ & $1.39\!\times\!10^{-2}$ & $1.14\!\times\!10^{-2}$ & $1.14\!\times\!10^{-2}$ & $0.01$ & $0.981$ \\
\textbf{Form 2 (Gated)} & \textbf{$5.94\!\times\!10^{-5}$} & \textbf{$1.10\!\times\!10^{-2}$} & \textbf{$9.01\!\times\!10^{-3}$} & \textbf{$9.01\!\times\!10^{-3}$} & $0.04$ & $0.960$ \\
Form 3 (G+F)            & $8.77\!\times\!10^{-5}$ & $1.36\!\times\!10^{-2}$ & $1.14\!\times\!10^{-2}$ & $1.14\!\times\!10^{-2}$ & $0.00$ & $0.989$ \\
\midrule
\dcpt{} (transfer)      & $5.73\!\times\!10^{-4}$ & $4.08\!\times\!10^{-2}$ & $3.91\!\times\!10^{-2}$ & $3.91\!\times\!10^{-2}$ & $0.04$ & $0.932$ \\
\bottomrule
\end{tabular}
\end{center}

\section*{Appendix B: Oracle baseline}
For reference, a \emph{\ptpp-wise oracle} that fits \dcpt{} directly on \ptpp{}=279 and evaluates on the same:
\begin{center}
\small
\begin{tabular}{lcccccc}
\toprule
Formulation & Huber$_{\log}$ & RMSE$_{\log}$ & MAE$_{\text{rel}}$ & MAPE$_{\text{clip}}$ & Interc. & Slope \\
\midrule
\dcpt{} (French)         & $2.05\!\times\!10^{-6}$ & $2.03\!\times\!10^{-3}$ & $1.63\!\times\!10^{-3}$ & $1.63\!\times\!10^{-3}$ & $0.00$ & $1.000$ \\
\dcpt{} (English/Arabic) & $1.67\!\times\!10^{-5}$ & $5.77\!\times\!10^{-3}$ & $4.44\!\times\!10^{-3}$ & $4.44\!\times\!10^{-3}$ & $-0.00$ & $1.001$ \\
\bottomrule
\end{tabular}
\end{center}
The Oracle uses full \ptpp{}=279 supervision and serves only as an upper bound.

\section*{Appendix C: In-sample grid (Form 3)}
\label{app:insample-grid}
\begin{figure}[h]
    \centering
    \includegraphics[width=\linewidth]{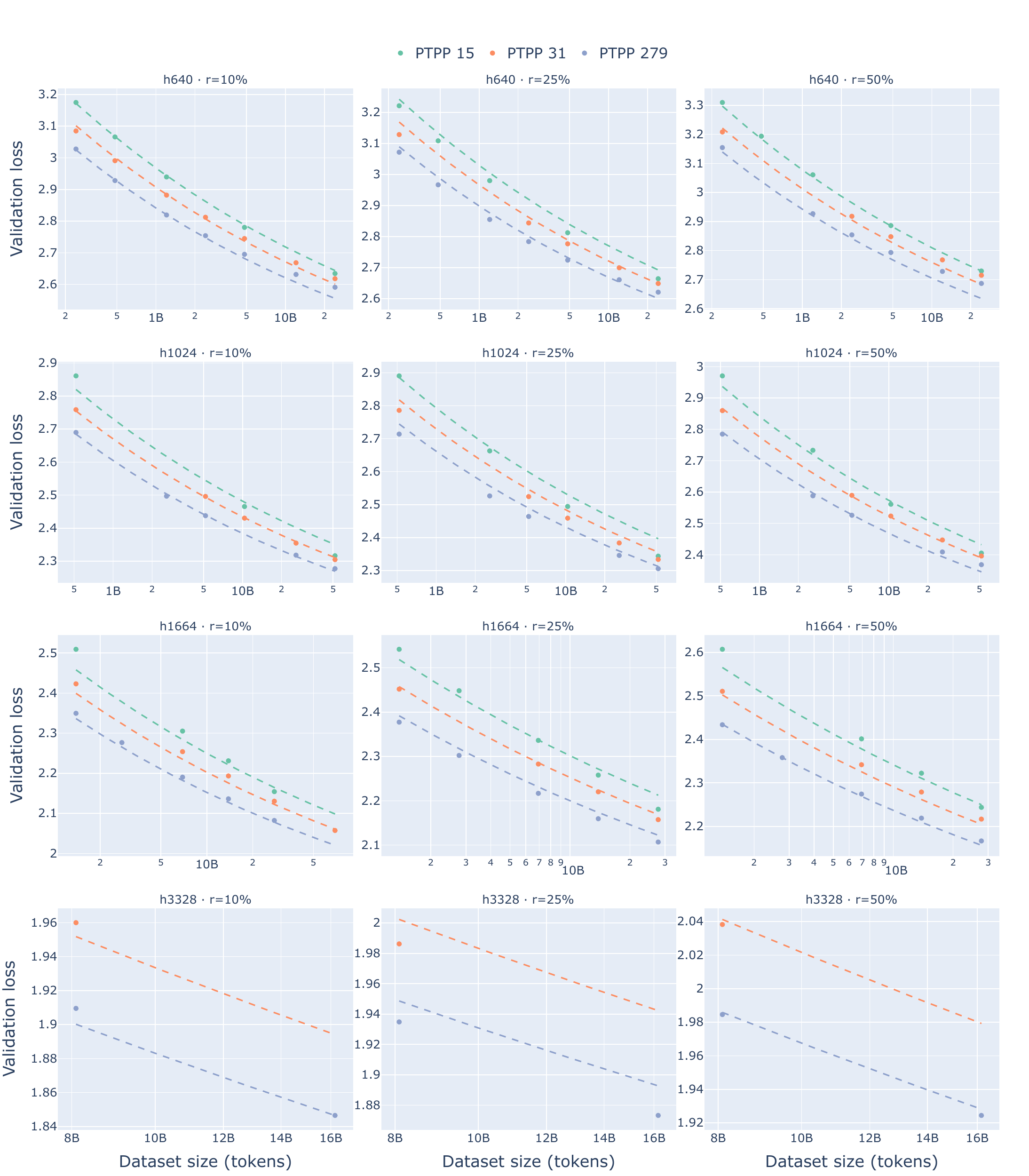}%
    \caption{In-sample fits for Form 3 (gated+floor). Rows: $r\in\{0.10,0.25,0.50\}$; columns: \{241M, 517M, 1.4B, 8.1B\}. Dashed: fitted curves; markers: observations. Used only as an auxiliary fit-quality check.}
    \label{fig:insample_grid}
\end{figure}

\end{document}